\documentclass[letterpaper, 10 pt, conference]{IEEEtran}

\IEEEoverridecommandlockouts 
\usepackage{graphicx}
\usepackage{siunitx}
\usepackage{amssymb,amsmath}
\usepackage{bm} 
\newtheorem{remark}{Remark}
\newtheorem{definition}{Definition}
\newtheorem{lemma}{Lemma}
\newtheorem{theorem}{Theorem}

\newtheorem{assumption}{Assumption}

\newtheorem{proofoftheorem}{Proof of Theorem}
\title{\LARGE \bf Asymptotic Allocation Rules for a Class of Dynamic Multi-armed Bandit Problems}

\author{T. W. U. Madhushani$^{1}$ and D. H. S. Maithripala$^2$ and N. E. Leonard$^{3}$% <-this % stops a space
	\thanks{$^{1}$Department of Mechanical and Aerospace Engineering, Princeton University, NJ 08544, USA.
		{\tt\small udarim@princeton.edu}}%
	\thanks{$^{2}$Department of Mechanical Engineering, University of Peradeniya, KY 20400, Sri Lanka.
		{\tt\small smaithri@pdn.ac.lk}}%
	\thanks{$^{3}$Department of Mechanical and Aerospace Engineering, Princeton University, NJ 08544, USA.
		{\tt\small naomi@princeton.edu}}%
}

\begin{document}

\maketitle
\thispagestyle{empty}
\pagestyle{empty}

\begin{abstract}
This paper presents a class of Dynamic Multi-Armed Bandit problems where the reward can be modeled as the noisy output of a time varying linear stochastic dynamic system that satisfies some boundedness constraints. The class allows many seemingly different problems with time varying option characteristics to be considered in a single framework. It also opens up the possibility of considering many new problems of practical importance. For instance it affords the simultaneous consideration of temporal option unavailabilities and the dependencies between options with time varying option characteristics in a seamless manner.
We show that, for this class of problems, the combination of any Upper Confidence Bound type algorithm with any efficient reward estimator for the expected reward ensures the logarithmic bounding of the expected cumulative regret. We demonstrate the versatility of the approach by the explicit consideration of a new example of practical interest. 
\end{abstract}
\allowdisplaybreaks

\section{Introduction} \label{sect:Introduction}
In decision theory Multi-Armed Bandit problems serve as a model that captures the salient features of human decision making strategies. The elementary case of a \textit{1-armed bandit} is a slot machine with one lever that results in a numerical reward  after every execution of the action. The reward is assumed to satisfy a specific but unknown probability distribution. A slot machine with multiple levers is known as a \textit{Multi-Armed Bandit} (MAB) \cite{Sutton,Robbins}. The problem is analogous to a scenario where an agent is repeatedly faced with several different options and is expected to make suitable choices in such a way that the cumulative reward is maximized \cite{Gittins}. This is known to be equivalent to minimizing the expected cumulative regret \cite{LaiRobbins}. 

Over decades optimal strategies have been developed to realize the above stated objective. In the standard multi-armed bandit problem the reward distributions are stationary. Thus if the mean values of all the options are known to the agent, in order to maximize the cumulative reward, the agent only has to sample from the option with the maximum mean. In reality this information is not available and the agent should choose options to maximize the cumulative reward while gaining sufficient information to estimate the true mean values of the option reward distributions. This is called the exploration-exploitation dilemma. In a case where the agent is faced with these choices with an infinite time horizon exploitation-exploration sampling rules are guaranteed to converge to the optimum option. In their seminal work Lai and Robbins \cite{LaiRobbins} established a lower bound for the cumulative regret for the finite time horizon case. Specifically, they establish a logarithmic lower bound for the number of times a sub-optimal option needs to be sampled by an optimal sampling rule if the total number of times the sub-optimal arms are sampled satisfies a certain boundedness condition.  
%%%%%%%%%%%%%%%%%%%%%%%%%%%%%%%%%%%%%%%%%%%%
The pioneering work by \cite{LaiRobbins} establishes a confidence bound and a sampling rule to achieve logarithmic cumulative regret. These results are further simplified in \cite{AgrawalSimpl} by establishing a confidence bound using a sample mean based method.  
Improving on these results, a family of Upper Confidence Bound (UCB) algorithms for achieving asymptotic and uniform logarithmic cumulative regret was proposed in \cite{Auer}. These algorithms are based on the notion that the desired goal of achieving logarithmic cumulative regret is realized by choosing an appropriate uncertainty model, which results in optimal trade-off between reward gain and information gain through uncertainty. 

What all these schemes have in common is a three step process: 1) a predication step, that involves the estimation of the expected reward characteristics for each option based on the information of the obtained rewards, 2) an objective function  that captures the tradeoff between estimated reward expectation and the uncertainty associated with it and 3) a decision making step that involves formulation of an action execution rule to realize a specified goal. For the standard MAB problem the reward associated with an option is considered as an iid stochastic process. Therefore in the frequentist setting the natural way of estimating the expectation of the reward is to consider the sample average \cite{LaiRobbins,AgrawalSimpl,Auer}. The papers \cite{Kauffman,Reverdy} present how to incorporate prior knowledge about reward expectation in the estimation step by leveraging the theory of conditional expectation in the Bayesian setting. We highlight that all these estimators ensure certain asymptotic bounds on the tail probabilities of the estimate of the expected reward. We will call such an estimator an \textit{efficient reward estimator}.
Furthermore all these methods with the exception of \cite{LaiRobbins} rely on UCB type algorithms for the decision making process. An extension to the standard MAB problem is provided in \cite{Kleinberg2010} to include temporal option unavailabilities where they propose a UCB based algorithm that ensures that the expected regret is upper bounded by a function that grows as the square root of the number of time steps.

In all of the previously discussed papers, the option characteristics are assumed to be static. However many real world problems can be modeled as multi-armed bandit problems with dynamic option characteristics  \cite{dacosta2008adaptive,Slivkins,granmo2010solving,Garivier2011,srivastava2014surveillance,schulz2015learning,tekin2010online}. In these problems reward distributions can change deterministically or stochastically. The work \cite{dacosta2008adaptive,Garivier2011,srivastava2014surveillance} present allocation rules and associated regret bounds for a class of problems where the reward distributions change deterministically after an unknown number of time steps. The paper  \cite{dacosta2008adaptive} presents a UCB1 based algorithm where they incorporate the Page-Hinkley change point detection method to identify the the point at which the underlying option characteristics change. A discounted UCB or a sliding-window UCB algorithm is proposed in \cite{Garivier2011} to solve non stationary MAB problems where the expectation of the reward switches to unknown constants at unknown time points. This work is extended in \cite{srivastava2014surveillance} by proposing sliding window UCL (SW-UCL) algorithm with adaptive window sizes for correlated Gaussian reward distributions. They incorporate the Page-Hinkley change point detection method to adjust the window size by identifying abrupt changes in the reward mean. Similarly, they also propose a block SW-UCL algorithm to restrict the transitions among arms.

A class of MAB problems with gradually changing reward distributions are considered in \cite{Slivkins,granmo2010solving}. Specifically 
\cite{Slivkins} considered the case where the expectation of the reward follows a random walk while \cite{granmo2010solving} addresses the problem where, at each time step, the expectation of each reward is modified by an independent Gaussian perturbation of constant variance. 
In \cite{schulz2015learning} the expectation of the reward associated with an option is considered to depend on a linear static function of some known variables that characterize the option and propose to estimate the reward based on learning this function.
A different class of dynamically and stochastically varying option characteristics is considered in \cite{tekin2010online} where the reward distribution of each option is modeled as a finite state irreducible, aperiodic, and reversible Markov chain.  

In this paper we consider a class of \textit{Dynamic Multi-Armed Bandit} problems (DMAB) that will include most of the previously stated dynamic problems as special cases. Specifically we consider a class of DMAB problems where the reward of each option is the noisy output of a multivariate linear time varying stochastic dynamic system that satisfies some boundedness conditions. This formulation allows one to accommodate a wide class of real world problems such as the cases where the option characteristics vary periodically, aperiodically, or gradually in a stochastic way. Furthermore incorporating this dynamic structure allows one to easily capture the underlying characteristic variations of each option as well as allow the possibility of incorporating dependencies between options. To the best of our knowledge this is the first time that such a wide class of dynamic problems have been considered in one general setting. We also incorporate temporal option unavailabilities into our structure that helps broaden the applicability of this model in real world problems. To the best of our knowledge it is the first time that temporal option unavailabilities are incorporated in a setting where the reward distributions are non-stationary. 

One major advantage of this linear dynamic systems formulation is that it immediately allows us to use the vast body of linear dynamic systems theory including that of switched systems to the problem of classification and solution of different DMAB problems. In this paper we prove that if the system characteristics satisfy certain boundedness conditions and the number of times the optimal arm becomes unavailable is at most logarithmic, then the expected cumulative regret is logarithmically bounded from above when one combines any UCB type decision making algorithm with any efficient reward estimator. We demonstrate the effectiveness of the scheme using an example where an agent intends to maximize the information she gathers under the constraint of option unavailability and periodically varying option characteristics.

In section-\ref{Secn:DMAB} we formally state the class of DMAB problems that is considered in this paper. 
We show in section-\ref{Secn:AsymptoticAllocationRules} that the combination of any UCB type allocation rule with an efficient estimator guarantees that the expected cumulative regret is bounded above by a logarithmic function of the number of time steps. In section-\ref{Secn:EfficientEstimators} we explicitly show, using a Hoeffding type tail bound \cite{Garivier2011}, that the sample mean estimator is an efficient estimator. Finally in section-\ref{Secn:Example} we provide a novel DMAB example that deals with unknown periodically and continuously varying options characteristics.
%%%%%%%%%%%%%%%%%%%%%%%%%%%%%%%%%%%%%%%%%%%

%%%%%%%%%%%%%%%%%%%%%%%%%%%%%%%%%%%%%%%
\section{Dynamic  Multi-Armed Bandit Problem}\label{Secn:DMAB}
In this paper we consider a wide class of dynamic multi-armed bandit problems where the reward is a noisy measurement of a linear time varying stochastic dynamic process.  The `noise' in the measurement and the `noise' in the process are assumed to have a bounded support. This is a reasonable and valid assumption since the rewards in physical problems are bounded and are greater than zero. Consider a \textit{k-armed bandit}. 
Let the reward associated with each option $i \in \{1,2,3,\ldots,k\}$ at the $t^{\mathrm{th}}$ time step be  given by the real valued random variable $X_i^t$. The expectation of this reward depends linearly on a $\mathbb{R}^m$ valued random variable $\theta^t$. 
The random variable $\theta^t$ represents option characteristics. The dynamics of the option characteristics can be multidimensional and thus we allow provision for $m$ to be larger than $k$. These option characteristics could either evolve deterministically or stochastically. The reward is assumed to depend linearly on the option characteristics. The dependence of the reward on the option characteristics may be precisely known or there could be some uncertainty about it. We model this uncertainty by an additive `noise' term with finite support.  We also allow the possibility of incorporating option dependencies and thereby considering the possibility of other options directly or indirectly influencing the reward associated with a given option. In order to capture this behavior in a concrete theoretical setting we assume that the bounded random variables $\theta^t \in \chi_{\theta}\subset \mathbb{R}^m$, with $\chi_{\theta}$ compact, and $X_i^t\in [0,\chi_x]$ with $0\leq \chi_x<\infty$, specifically satisfy a linear time varying stochastic process, 
\begin{align}
\theta^t&=A^t\theta^{t-1}+B^tn^t_{\theta},\label{eq:Process}\\
X_i^t&=\gamma_i^t\,\left(H_i^t\theta^t+g_i^{t}\,n^t_{xi}\right),\label{eq:NoisyReward}
\end{align} 
where $\{n^t_{\theta}\}$ is a bounded $\mathbb{R}^q$ valued stochastic process with zero mean and constant covariance $\Sigma_\theta$ while $\{n^t_{xi}\}$ is a $\mathbb{R}$ valued bounded stochastic process with zero mean and constant variance $\sigma_{xi}$. We also let $\{\gamma_i^t\},\{g_i^t\}$ be real valued deterministically varying sequences while $\{A^t\},\{B^t\},\{H_i^t\}$ are matrix valued deterministic sequences of appropriate dimensions. We allow the variances, $\sigma_{xi}^2$, corresponding to each arm to be different. Letting $\gamma_i^t\in \{0,1\}$ allows us to consider temporal option unavailabilities. 

Expression-(\ref{eq:Process}) describes the collective time varying characteristics of all the options and the absence or presence of $B^tn_{\theta}$ dictates whether these dynamics are deterministic or stochastic. 
Expression-(\ref{eq:NoisyReward}) describes how the reward depends on the option characteristics. The presence of the `noise' term $g_i^{t}\,n_{xi}$ indicates that the rewards that one obtains given the knowledge of the option characteristics involve some bounded uncertainty. The case where $\{A^t\},\{B^t\},\{H_i^t\}$ each has a block diagonal structure represents independent arms and the case where there are off diagonal entries represent situations where the arms depend on each other.
Notice that by setting $A^t\equiv I$ and $B^t\equiv 0$ we obtain the standard MAB with temporal option unavailabilities. One major advantage of this linear dynamic systems formulation is that it allows one to use the vast body of linear dynamic systems theory including that of switched systems in the classification and solution of different DMAB problems.

From equations (\ref{eq:Process}) and (\ref{eq:NoisyReward}) we see that the expectations $E(\theta^t),E(X_i^t)$ evolve according to
\begin{align}
E(\theta^t)&=A^tE(\theta^{t-1}),\label{eq:EProcess}\\
E(X_i^t)&=\gamma_i^t\,H_i^tE(\theta^t),\label{eq:EReward}
\end{align}
and that the covariances $\Sigma(\theta^t)\triangleq E(\theta^{t}{\theta^{t}}^T)-E(\theta^t){E(\theta^t)}^T$, $\Sigma(X_i^t)\triangleq E(X_i^{t}{X_i^{t}}^T)-E(X_i^t){E(X_i^t)}^T$ evolve according to
\begin{align}
\Sigma(\theta^t)&=A^t\Sigma(\theta^{t-1}){A^t}^T+B^t\Sigma_\theta{B^t}^T,\label{eq:VProcess}\\
\Sigma(X_i^t)&=(\gamma_i^t)^2\,\left(H_i^t\Sigma(\theta^t){H_i^t}^T+\sigma_{xi}^2{(g_i^t)}^2\right).\label{eq:VReward}
\end{align}
Boundedness of $\theta_i^t$ implies $E(\theta^{t})$ and $\Sigma(\theta^t)$ should remain bounded. 

Let $\Phi^t_\tau\triangleq \left(\prod_{j=\tau}^tA^j\right)$ and then since $E(\theta^t)=\Phi^t_1E(\theta^0)$ we find that the expectation and the covariance of the reward become unbounded if $\lim_{t\to\infty}||\Phi^t_1||=\infty$. On the other hand the expectation converges to zero if $\lim_{t\to\infty}||\Phi^t_1||=0$. Thus sequences $\{A^t\}$ that satisfy the conditions $\limsup_{t\to\infty}||\Phi^t_1||=\bar{a}<\infty$ and $\liminf_{t\to\infty}||\Phi^t_1||={a}>0$ are the only ones that correspond to a meaningful DMAB problem.  Thus to ensure boundedness of $E(\theta^{t})$ we assume that:
\begin{assumption}\label{As:MainAssumption0}
The sequence $\{A^t\}$ satisfies:
\begin{align}
\limsup_{t\to\infty}\left|\left|\prod_{j=1}^tA^j\right|\right|&<\infty\\
\liminf_{t\to\infty}\left|\left|\prod_{j=1}^tA^j\right|\right|&>0
\end{align}
and $\exists \:\: a,\bar{a}>0$ such that,
\begin{align}
a<\left|\left|\prod_{j=\tau}^tA^j\right|\right|<\bar{a},\:\:\:\:\:
\end{align}
$\forall \:\: t\geq\tau$. 
\end{assumption}

Several examples of sequences $\{A^t\}$ of practical significance that ensure this condition are those where $A^t$: 
\begin{enumerate}
\item is an orthogonal matrix or is a stochastic matrix (i.e. $||A^t||=1$),
\item is a periodic matrix (i.e. $A^t=A^{t+N}$ for some $N>0$),
\item corresponds to a stable switched system.
\end{enumerate}

{Next we will consider conditions needed for the boundedness of $\Sigma(X_i^t)$. Note that the covariance of $\theta^t$ is given by
{\small
\begin{align}
\Sigma(\theta^t)&=\Phi^t_1\Sigma(\theta^{0}){\Phi^t_1}^T+\sum_{\tau=1}^t\Phi^{t}_{\tau+1}B^{\tau}\Sigma_\theta{B^{\tau}}^T(\Phi^{t}_{\tau+1})^T.\label{eq:VProcess1}
\end{align}
}
Assumption-\ref{As:MainAssumption0} ensures that the first term on the right hand side is bounded and that
{\small
\begin{align}
||\Sigma(\theta^t)||&\leq \bar{a}^2||\Sigma(\theta^{0})||+\bar{a}^2||\Sigma_\theta||^2\sum_{\tau=1}^t||B^\tau||^2.\label{eq:VProcess2}
\end{align}
}
Thus it also follows that $||\Sigma(\theta^t)||$ remains bounded in any finite time horizon if $||B^t||$ remains bounded in that period. On the other hand
if the sequence $\{B^t\}$ satisfies $||B^t||\leq c/t$ for some $c>0$ or if the number of time steps where the condition $\Phi^{t}_{\tau}B^{\tau-1}\neq 0$ is satisfied remains finite then $||\Sigma(\theta^t)||$ is guaranteed to be bounded for all $t>0$. Therefore from (\ref{eq:VReward}) we find that in order to satisfy the boundedness of $X_i^t$ the sequences $\{\gamma_i^t\},\{||H_i^t||\},\{g_i^t\}$ must necessarily be bounded from above in addition to what is specified in Assumption \ref{As:MainAssumption0}. 

In order to define a meaningful DMAB problem the notion of an optimal option should be well defined. 
That is $i^*\triangleq \arg\limits_{i}\max\{H_i^tE(\theta^t)\}$ is independent of time. The following assumption specifies the conditions necessary for the boundedness of the reward $X_i^t$ as well as the conditions necessary for the existence of an optimal arm.
\begin{assumption}\label{As:MainAssumption}
We will assume that the sequences $\{\gamma_i^t\},\{B^t\},\{H_i^t\},\{g_i^t\}$ guarantee the following conditions for all $t>0$: 
\begin{align}
||\Sigma(\theta^t)||\leq &\sigma,\label{eq:SigmaBnd}\\
\gamma_i^t\in \{0,1\}&,\\
0<g_i^t\leq&\bar{g}_i,\\
||B^t|| \leq \frac{b}{t}\:\:\:\mbox{or}\:\: &||B^t||\neq 0 \:\:\:\mbox{finitely many times},\\
h_i<||H_i^t||\leq&\bar{h}_i,\label{eq:HBnd}
\end{align}
and $\forall \: t\geq 0$ there exists a unique $i^*=i^t_*$ such that
\begin{align}
\Delta_{i}\leq \Delta_i^t&\triangleq {H_{i^t_*}}^tE(\theta^t_{i^t_*})-H_i^t E(\theta^t) \leq  \bar{\Delta},\label{eq:OptimalArm}
\end{align}
$\forall \:\:i\neq i^t_*$ and 
while
\begin{align}
\sum_{j=2}^t\mathbb{I}_{\{\gamma^j_{i^*}=0\}} \leq \gamma \log{t},\label{eq:LogBndAvailability}
\end{align}
for some $\bar{g}_i,h_i,\bar{h}_i,\bar{\Delta},{\Delta}_i,\gamma,\sigma,b>0$ where $\mathbb{I}_{\{\gamma^j=0\}}$ is the indicator function.
\end{assumption}

Note that condition (\ref{eq:OptimalArm}), which implies existence of a well defined optimal arm, is guaranteed if
$({h_{i^*}}a||E(\theta^0_{i^*})||-\bar{h}_i\bar{a}||E(\theta^0)||)>0, \:\: \forall \:\: t\geq \tau>0$. Condition (\ref{eq:LogBndAvailability}) implies that this optimal arm becomes unavailable at most logarithmically with the number of time steps. Finally the boundedness of $X_i^t$ is guaranteed by the conditions (\ref{eq:SigmaBnd}) -- (\ref{eq:HBnd}).}

%%%%%%%%%%%%%%%%%%%%%

We will now proceed to analyze the regret of the DMAB problem stated above. Consider the probability space $(\Omega,\mathcal{U},\mathcal{P})$ and the increasing sequence of subalgebras $\mathcal{F}_{0}\subset\mathcal{F}_{1}\cdots \subset\mathcal{F}_{t}\cdots \subset\mathcal{F}_{n-1}\subset \mathcal{U}$ for $t=0,1,\cdots,n$ where 
$\mathcal{P}$ is the probability measure on the sigma algebra $\mathcal{U}$ of $\Omega$. The sigma algebra $\mathcal{F}_{t}$ represents the information that is available at the $t^{\mathrm{th}}$ time step.
Let $\{\varphi_t\}_{t=1}^n$ be a sequence of random variables, each defined on $(\Omega,\mathcal{F}_{t-1},\mathcal{P})$ and taking values in $\{1,2,\cdots,k\}$. The random variable $\varphi_t$ models the action taken by the agent at the $t^{\mathrm{th}}$ time step. The value $i\in\{1,2,\cdots,k\}$ of the random variable $\varphi_t$ specifies that the  $i^{\mathrm{th}}$ option is chosen at time step $t$. Then $\mathbb{I}_{\{\varphi_t =i\}}$ is the $\mathcal{F}_{t-1}$ measurable indicator random variable that takes a value one if the $i^{\mathrm{th}}$ option is chosen at step $t$ and is zero otherwise.

The DMAB problem is to find an allocation rule $\{\varphi_t\}_{t=1}^n$ that maximizes the expected cumulative reward or equivalently that minimizes the cumulative regret.  
The cumulative reward after the the $n^{\mathrm{th}}$ time step is defined to be the real valued random variable $S_n$ defined on the probability space $(\Omega,\mathcal{F}_{n-1},\mathcal{P})$ that is given by
 \begin{align*}
 S_n&=\sum_{t=1}^n\sum_{i=1}^k E(X_i^t\mathbb{I}_{\{\varphi_{t} =i\}}|\mathcal{F}_{t-1})\\
&=\sum_{t=1}^n\sum_{i=1}^k E(X_i^t|\mathcal{F}_{t-1})\mathbb{I}_{\{\varphi_{t} =i\}}.
 \end{align*}
 Thus the expected cumulative reward is,
 \begin{align*}
 E(S_n)=\sum_{t=1}^n\sum_{i=1}^k E({X}_i^t)E(\mathbb{I}_{\{\varphi_{t} =i\}})
 \end{align*}
 where $T_i(n)=\sum_{t=1}^n\mathbb{I}_{\{\varphi_t =i\}}$ is a real valued random variable defined on $(\Omega,\mathcal{F}_{n-1},\mathcal{P})$ that represents the number of times the $i^{\mathrm{th}}$ arm has been sampled in $n$ trials. Note that 
 $E(X_i^t)=\gamma_i^tH_i^tE(\theta^t)$.

Let $i^t_*=\max_i\{E(X_i^t)\}$. Then the expected cumulative regret is defined as
{\small
 \begin{align}
R_n&\triangleq \sum_{t=1}^n\left(E({X}^t_{i_t^*})- \sum_{i=1}^k\gamma_{i}^tE({X}_i^t)E(\mathbb{I}_{\{\varphi_{t} =i\}}) \right).
\label{eq:DynRegret}
 \end{align} 
 }
Then from condition (\ref{eq:OptimalArm}) we find that
{\small
 \begin{align*}
R_n
 &=\sum_{i=1}^k\sum_{t=1}^n\mathbb{I}_{\{\gamma_{i^*}^t=1\}}\left(H_{i^*}^tE(\theta_{i^*}^t)- \gamma_i^tH_{i}^tE(\theta_{i}^t) \right)E(\mathbb{I}_{\{\varphi_{t} =i\}})\nonumber\\
 &+\sum_{i=1}^k\sum_{t=1}^n\mathbb{I}_{\{\gamma_{i^*}^t=0\}}\left(H_{i^t_*}^tE(\theta_{i^t_*}^t)- \gamma_i^tH_{i}^tE(\theta_{i}^t) \right)E(\mathbb{I}_{\{\varphi_{t} =i\}})\nonumber\\
 & \leq \bar{\Delta}\sum_{i\neq i^*}^kE\left(T_i(n)\right).
 \end{align*} 
 }
In their seminal work \cite{LaiRobbins} Lai and Robbins proved that, for the static MAB problem, the regret is bounded below by a logarithmic function of the number of time steps.

%%%%%%%%%%%%%%%%%%%%%%%%%%%%%%%
\section{Asymptotic Allocation Rules for the DMAB Problem}\label{Secn:AsymptoticAllocationRules}
In this section we show how to construct \textit{asymptotically efficient} allocation rules for the class of DMAB problems that were formally defined above. Specifically, in the following, we will show that the combination of any \textit{UCB based} decision making process and an \textit{efficient estimator} provides such an allocation rule.

In the DMAB problem $\mu_i^t\triangleq E(X_i^t)$ is time varying. Thus one needs to consider a `time average' for $\mu_i^t$. This time average depends on how one samples option $i$. Specifically it is a $\mathcal{F}_{t-1}$ measurable random variable
\begin{align}
\widehat{\mu}_i^t&\triangleq \frac{1}{T_i(t)}\sum_{j=1}^{t}E(X_i^j)\mathbb{I}_{\{\psi_j=i\}}.
\label{eq:TimeAverageMean}
\end{align}
This random variable can not be estimated using the maximum likelihood principle since $E(X_i^j)$ are unknown and thus will have to be estimated by other means. We will consider a $\mathcal{F}_{t-1}$ measurable random variable $\widehat{X}_i^t$ to be an estimator of $\widehat{\mu}_i^t$ if 
$E(\widehat{X}_i^t)=E(\widehat{\mu}_i^t)$.

\begin{definition}
Let $\widehat{X}_i^t$ be a $\mathcal{F}_{t-1}$ measurable random variable such that
$E(\widehat{X}_i^t)=E(\widehat{\mu}_i^t)$ and $T_i(t)$ be the $\mathcal{F}_{t-1}$ measurable random variable that represents the number of times the
$i^{\mathrm{th}}$ option has been sampled up to time $t$. An estimator $\widehat{X}_i^t$ that ensures
\begin{align}
	\mathcal{P}\left(\widehat{X}_i^t\geq \widehat{\mu}_i^t+\sqrt{\frac{\vartheta}{T_i(t)}}\right)&\leq \frac{\nu\,\log{t}}{\exp\left(2\kappa \vartheta\right)},\label{eq:TailProbBnd1}\\
	\mathcal{P}\left(\widehat{X}_i^t\leq \widehat{\mu}_i^t-\sqrt{\frac{\vartheta}{T_i(t)}}\right)&\leq \frac{\nu\,\log{t}}{\exp\left(2\kappa \vartheta\right)}. \label{eq:TailProbBnd2}
\end{align}
for some $\kappa,\vartheta,\nu>0$ will be referred to as an 
\emph{efficient reward estimator}.
\end{definition}

In section-\ref{Secn:EfficientEstimators} we show that the frequentist average mean estimator satisfies this requirement.

\begin{definition}
Let $\widehat{X}_i^t$ be a $\mathcal{F}_{t-1}$ measurable random variable such that
$E(\widehat{X}_i^t)=E(\widehat{\mu}_i^t)$. The allocation rule $\{\varphi_{t}\}_1^{n}$ will be referred to as
\emph{UCB based}  if it is chosen such that
	\begin{align}
	\mathbb{I}_{\{\varphi_{t+1}=i\}}=\left\{
	\begin{array}{cl} 1 & \:\:\:Q^{t}_i=\max\{Q^{t}_1,\cdots,Q^{t}_k\}\label{eq:UCBallocation}\\
  	0 & \:\:\: {\mathrm{o.w.}}\end{array}\right.
	\end{align}
	with 	
	\begin{align}
	Q^t_i&\triangleq \widehat{X}^t_i+\sigma\sqrt{\frac{\Psi\left(t\right)}{T_i(t)}}\label{eq:UCBQ}
	\end{align}
where $\Psi(t)$ is an increasing function of $t$ with $\Psi(1)=0$ and $\sigma>0$. We will also let $T_i(1)=1$ for all $i$. 
\end{definition}

\begin{remark}
There exists two choices for  picking an option at the first time step. If there exists some prior knowledge one can use that as the initial estimate $\widehat{X}^1_i$. On the other hand in the absence of such prior knowledge one can sample every option once and select the the sampled values for $\widehat{X}^1_i$.
\end{remark}

We will show below that combining any efficient estimator with a UCB based allocation rule ensures that the number of times a suboptimal arm is sampled is bounded above by a logarithmic function of the number of samples. This result is formally stated below and is proved in the appendix.
\begin{theorem}\label{Theom:DynamicRegret}
Let conditions specified in Assumption \ref{As:MainAssumption0} and \ref{As:MainAssumption} hold.
Then any efficient estimator combined with an UCB based allocation rule $\{\varphi_t\}_1^{n}$ ensures that for every $i\in\{1,2,\cdots,k\}$ such that $i\neq i^{*}$ and for some $ l\geq 1$.
\begin{align*}
E(T_i(n))\leq  \gamma\log{n}+ \frac{4\sigma^2}{{\Delta^2_i}}\,\Psi(n) +\left(l+\nu\sum_{t=l}^{n-1}\frac{\log{t}}{t^{2\kappa\sigma^2\alpha}}\right).
\end{align*}
Thus  the cumulative expected regret satisfies:
\begin{align*}
R_n\leq c_0\log{n} + c_1,
\end{align*}
for some constant $c_0,c_1>0$ if $\Psi(t)$ satisfies
\begin{align*}
\alpha \log{t}\leq \Psi(t) \leq \beta \log{t},
\end{align*}
for some constants $3/(2\kappa\sigma^2)<\alpha\leq \beta$.
\end{theorem}

\begin{remark}
If one selects $\Psi(t)=16\log{t}$ one obtains the standard UCB-Normal algorithm proposed in \cite{Auer} while if one selects ${\Psi(t)}=\left(\Phi^{-1}\left(1-1/(\sqrt{2\pi e}\,t^2)\right)\right)^2$ where $\Phi^{-1}\left(\cdot\right)$ is the inverse of the cumulative distribution function for the normal distribution one obtains the UCL algorithm proposed in \cite{Reverdy}. 
\end{remark}
%%%%%%%%%%%%%%%%%%%%%%%%%%%%%%
\subsection{Efficient Estimators}\label{Secn:EfficientEstimators}

Let $S_i^t$ be the $\mathcal{F}_{t-1}$ measurable random variable that gives the cumulative reward received by choosing arm $i$ up to the $t^{\mathrm{th}}$ time step that is given by
\begin{align*}
S_i^t=\sum_{j=1}^{t}E(X_i^j\mathbb{I}_{\{\psi_j=i\}}|\mathcal{F}_{j-1})
=\sum_{j=1}^{t}E(X_i^j|\mathcal{F}_{j-1})\mathbb{I}_{\{\psi_j=i\}}.
\end{align*}

Define the $\mathcal{F}_{t-1}$ measurable simple average sample mean estimate $\widehat{X}_i^t$ of the cumulative mean reward received from arm $i$ as 
\begin{align}
\widehat{X}_i^t&\triangleq \left\{\begin{array}{lc}\widehat{X}_i^1 & \mathrm{if}\:\: T_i(t)=0\\
\frac{S_i^t}{T_i(t)} & \mathrm{o.w.}
\end{array}\right..\label{eq:MeanAverage}
\end{align}
Then
\begin{align*}
E(\widehat{X}_i^t)&=\sum_{j=1}^{t}E\left(\frac{E(X_i^j| \mathcal{F}_{j-1})\mathbb{I}_{\{\psi_j=i\}}}{T_i(t)}\right).
\end{align*}
Since $E(X_i^j| \mathcal{F}_{j-1})$ is independent of $\mathbb{I}_{\{\psi_j=i\}}$ and $T_i(t)$ we have that
\begin{align*}
E(\widehat{X}_i^t)&=\sum_{j=1}^{t}E(X_i^j)E\left(\frac{\mathbb{I}_{\{\psi_j=i\}}}{T_i(t)}\right)\nonumber \\
&=E\left(\sum_{j=1}^{t}\frac{E(X_i^j)\mathbb{I}_{\{\psi_j=i\}}}{T_i(t)}\right)=E(\widehat{\mu}_i^t).
\end{align*}

The tail probability distribution for the above sample mean average is given by the following lemma  which follows from Theorem 4 of \cite{Garivier2011}.
\begin{lemma}
If the random process $\{{X}_i^t\}$ satisfies ${X}_i^t\in [0,\chi_x], \:\:\forall i\in{1,2,\ldots,k}$ and $t>0$ and $\widehat{X}_i^t,\widehat{\mu}_i^t$ are given by (\ref{eq:MeanAverage}) and (\ref{eq:TimeAverageMean}) respectively we have that,
\begin{align*}
\mathcal{P}\left(\widehat{X}_i^t>\widehat{\mu}_i^t+\sqrt{\frac{\vartheta}{T_i(t)}}\right)\leq  
\frac{\nu\,\log{t}}{\exp\left(2\kappa \vartheta\right)}
\end{align*}
where $\kappa=\left(1-\frac{\eta^2}{16}\right)/\chi^2$ and $\nu=1/\log (1+\eta)$ for all $t>0$ and $\eta,\vartheta>0$. 
\end{lemma}
%%%%%%%%%%%%%%%%%%%%%%%%%%%%%%
Thus we have that (\ref{eq:TailProbBnd1}) and (\ref{eq:TailProbBnd2}) are satisfied for the sample mean estimate of the reward.

%%%%%%%%%%%%%%%%%%%%%%%%%%%%
\section{Example: Periodically Continuously Varying Option Characteristics}\label{Secn:Example}
In this section we consider a novel example of practical interest.  
The problem that we consider is that of an agent trying to maximize the reward that depends continuously on certain, periodically and continuously varying, option characteristics. The agent is assumed to be unaware of any information about this periodic behavior.

%%%%%%%%%%%%%%%%%
Specifically we consider the problem where an agent is encountered with $k$ number of options. Each option may vary with time and may become unavailable from time to time. For this example we assume that the options do not depend on each other.
This is the case, for instance, if the agent is dealing with collecting human behavioral information in a recreational park and has several options for locating herself for the purpose of collecting this information. The average number of people who frequent the park may vary depending on whether it is in the morning, afternoon or evening. Similar circumstances occur if one needs to select the type of optimal crops, highlight a particular  product in a store, sample a set of sensors whose characteristics vary with the time of the day, or advertise a particular event in super markets. In each of these cases due to certain external events some of the options may become temporally unavailable as well.  This class is characterized by a fixed periodic block diagonal matrix $A^t=\mathrm{diag}(A^t_1,A^t_2,\cdots,A^t_k)$ where each $A_i^t$ is a $p\times p$ matrix that satisfies the property $A_i^{t+N}=A_i^t$ for some $N>1$. The matrix $A_i^t$ encodes how the dynamics of the expected characteristics of the $i^{\mathrm{th}}$ option varies. For the example we consider here the options do not depend on each other thus we will also set $B^t=[B_1,B_2,\cdots,B_k]^T$ where each $B_i$ is a $1\times p$ row matrix and each $H_i^t=H_i$ has a corresponding block structure so that the option dynamics are not coupled.

We will consider two cases. One where we will assume that the total number of people who visit a park is always the same for every day and the more realistic case where the number of people that visit a park varies stochastically. For the first case we will set $B^tn^t_{\theta}\equiv 0$ which guarantees the boundedness of the covariance of $\theta^t$ for infinite time horizons as well. In the second case we pick $B^tn^t_{\theta}\neq 0$ where $B^t\equiv B$ is a constant matrix and $n^t_{\theta}$ is a bounded random process with zero mean. In this second case the boundedness of the covariance of $\theta^t$ is guaranteed only for finite time horizons.

For illustration we use $k=5$ and $N=3$.
We see that this problem can be modeled by selecting 
$\theta^t=(\theta^t_1,\theta^t_2,\cdots,\theta^t_5)$ where each $\theta^t_i\in \mathbb{R}^3$. We let $A^t=\mathrm{diag}(A^t_1,A^t_2,\cdots,A^t_5)$ where each $A^t_i=(A_i)^t$ with
\begin{align*}
A_i=\begin{bmatrix} 0 & 1 & 0\\ 0 & 0 & 1\\ 1 & 0 & 0\end{bmatrix}.
\end{align*}
The output matrix $H^t_i$ is a constant $1\times 3k$ row matrix and has all zero entries with the exception of the $(3i-2)^{\mathrm{th}}$ entry which is equal to one. 
The direct noise coupling term corresponding to each reward is chosen to be $g^t_i\equiv 1$ and we will assume that the noise is a bounded random variable $n^t_{x_i}$ with zero mean.

Thus 
we see that the expected reward of the $i^{\mathrm{th}}$ option satisfies
$E(X_i^t)=H(A_i)^t\theta_i^0$ where $H=[1\:\:\:0\:\:\:0]$ and hence satisfies the condition $E(X_i^{t+3})=E(X_i^t)$ for all $t>0$.
The initial condition is chosen such that $\theta^0=(\theta^0_1,\theta^0_2,\cdots,\theta^0_5)$ where each $\theta_i^0$ takes the form $\theta_i^0=\bar{\theta}_i[\alpha_1\:\:\: \alpha_2\:\:\:\alpha_3]^T$ with $\bar{\theta}_i\in \mathbb{R}$.
The real positive constants $\alpha_1, \alpha_2, \alpha_3$ satisfy the condition
$\alpha_1\alpha_2\alpha_3=1$ and captures the periodic variation of the number of visitors within a given day. The block diagonal structure of $A^t,B^t,H_i^t$ amounts to the assumption that the number of people that frequent different locations are uncorrelated. 

The total number of people that visit the park in a particular day is given by
$(\bar{\theta}_1+\bar{\theta}_2+\bar{\theta}_3+\bar{\theta}_4+\bar{\theta}_5)(\alpha_1+\alpha_2+\alpha_3)$.
In the first case we will assume that each $\bar{\theta}_i$ is fixed for each day and thus set $B^t\equiv 0$. In the other case we will consider the more realistic situation where this value changes stochastically according to a uniform distribution on the support $[\bar{\theta}_i-50,\bar{\theta}_i+50]$. That is, we select $n_\theta^t$ to be uniformly distributed on $[-50,50]$. We also let the noise term $n_{xi}$ for each $i$ to be uniformly distributed on $[-50,50]$. Notice that in this second case the growth condition for the covariance $\Sigma(X_i^t)$ is only satisfied for a finite time horizon.

For the simulations we let $\alpha_1=3/4,\alpha_2=1,\alpha_3=4/3$ and $\bar{\theta}_1=400, \bar{\theta}_2=350, \bar{\theta}_3=750, \bar{\theta}_4=1000, \bar{\theta}_5=526$. Thus the optimal option is $i^*=4$ and is well defined for all $t$. 
For the simulations we use the UCB algorithem (\ref{eq:UCBallocation}) -- (\ref{eq:UCBQ}) with $\Psi(t)=16\log{t}$ and the standard frequentist average estimator (\ref{eq:MeanAverage})  for the estimation of the rewards. 
In compliance with the assumption that the optimal option becomes unavailable at most logarithmically, we let
\begin{align*}
\gamma_i^t&=\left\{\begin{array}{lc} 0 & \mathrm{if}\:\:\left([\log{(n_i+t+1)}]-[\log{(n_i+t)}]\right)=1\\
1 & \mathrm{o.w.}
\end{array}\right.,
\end{align*}
for some integer $n_i>0$ where $[x]$ denotes the nearest integer value of $x$. For convenience of simulation we will let $\gamma_i^t\equiv 1$ for $i\neq i^*$ and $\{\gamma_{i^*}^t\}$ as chosen above. 

We estimate the expected reward, $E(S_n)$, the expected cumulative regret, $E(R_n)$, and the expected number of times the optimal arm is selected, $E(T_{i^*}(n))$, by simulating the algorithm for each $1\leq n\leq200$ a $1000$ times and by computing the frequentist mean as an estimate for $E(S_n)$, $E(R_n)$, and  $E(T_{i^*}(n))$.

\begin{figure}[h!]
    \centering    \includegraphics[width=0.5\textwidth]{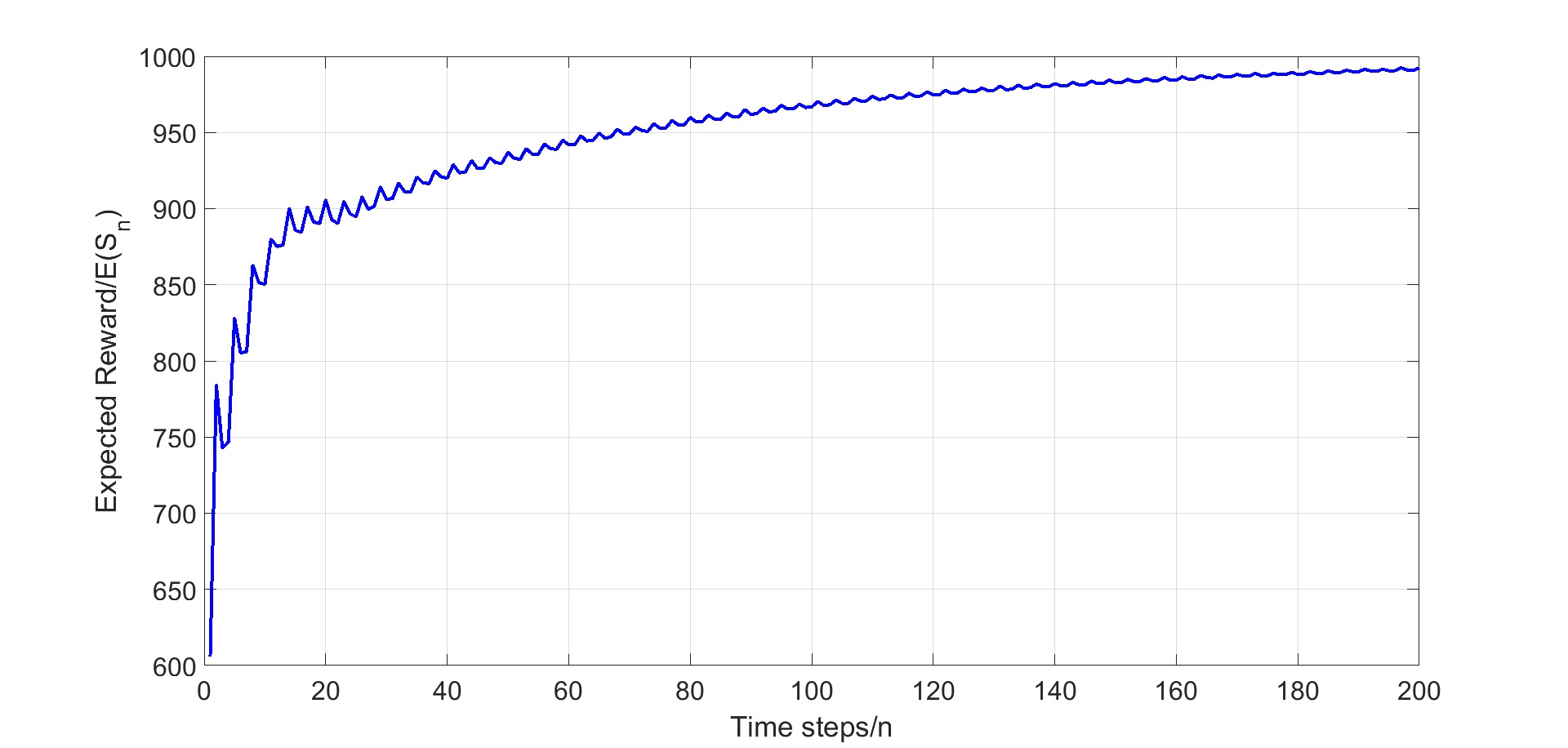}
 \caption{Expected reward $E(S_n)$ for the time varying DMAB with no process noise.}
\label{Fig:ErewardNoNoise}
\end{figure}

\begin{figure}[h!]
    \centering    \includegraphics[width=0.5\textwidth]{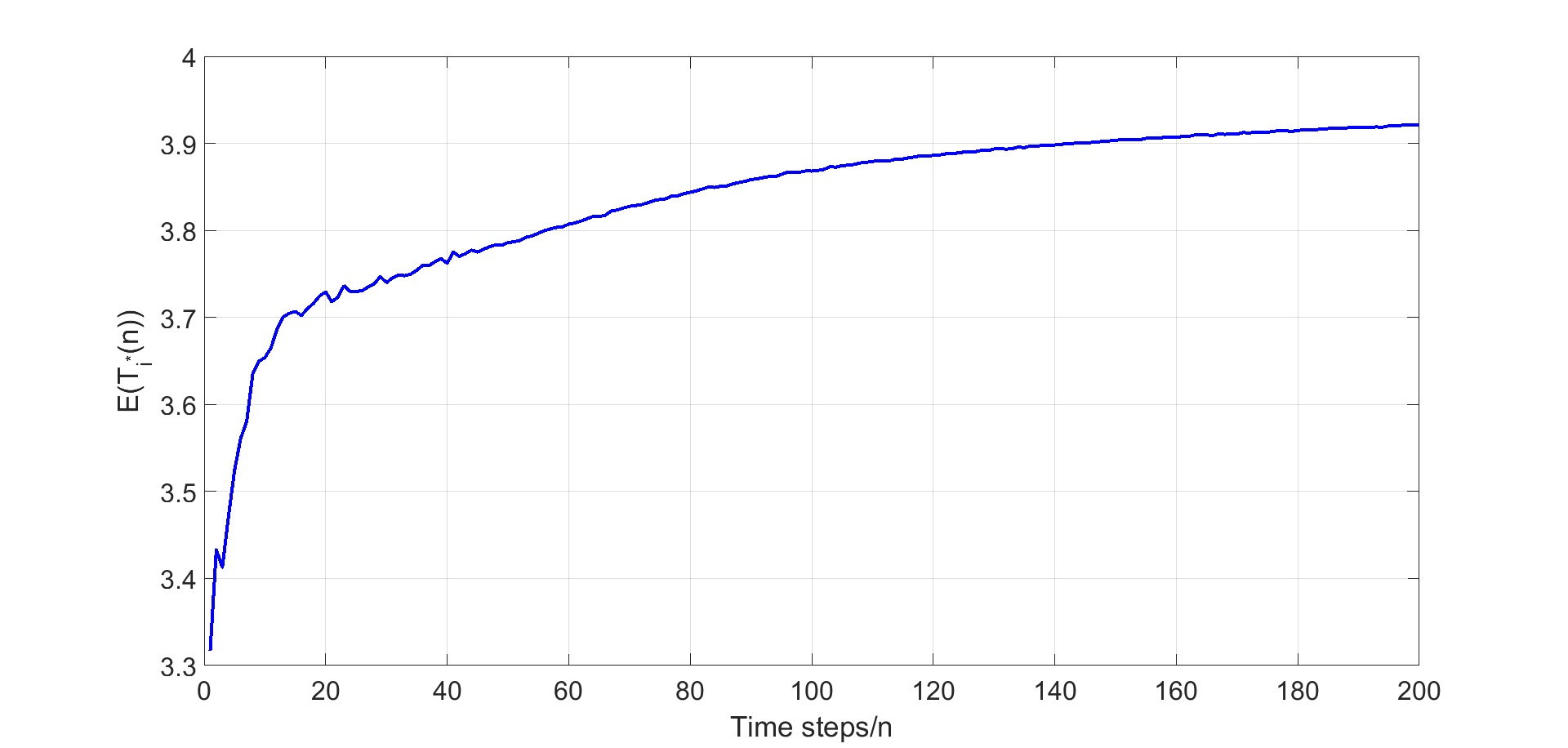}
 \caption{Expected number of times, $E(T_{i^*}(n))$, the optimal arm has been sampled  for the time varying DMAB with no noise. }
\label{Fig:EToptNoNoise}
\end{figure}

\begin{figure}[h!]
    \centering    \includegraphics[width=0.5\textwidth]{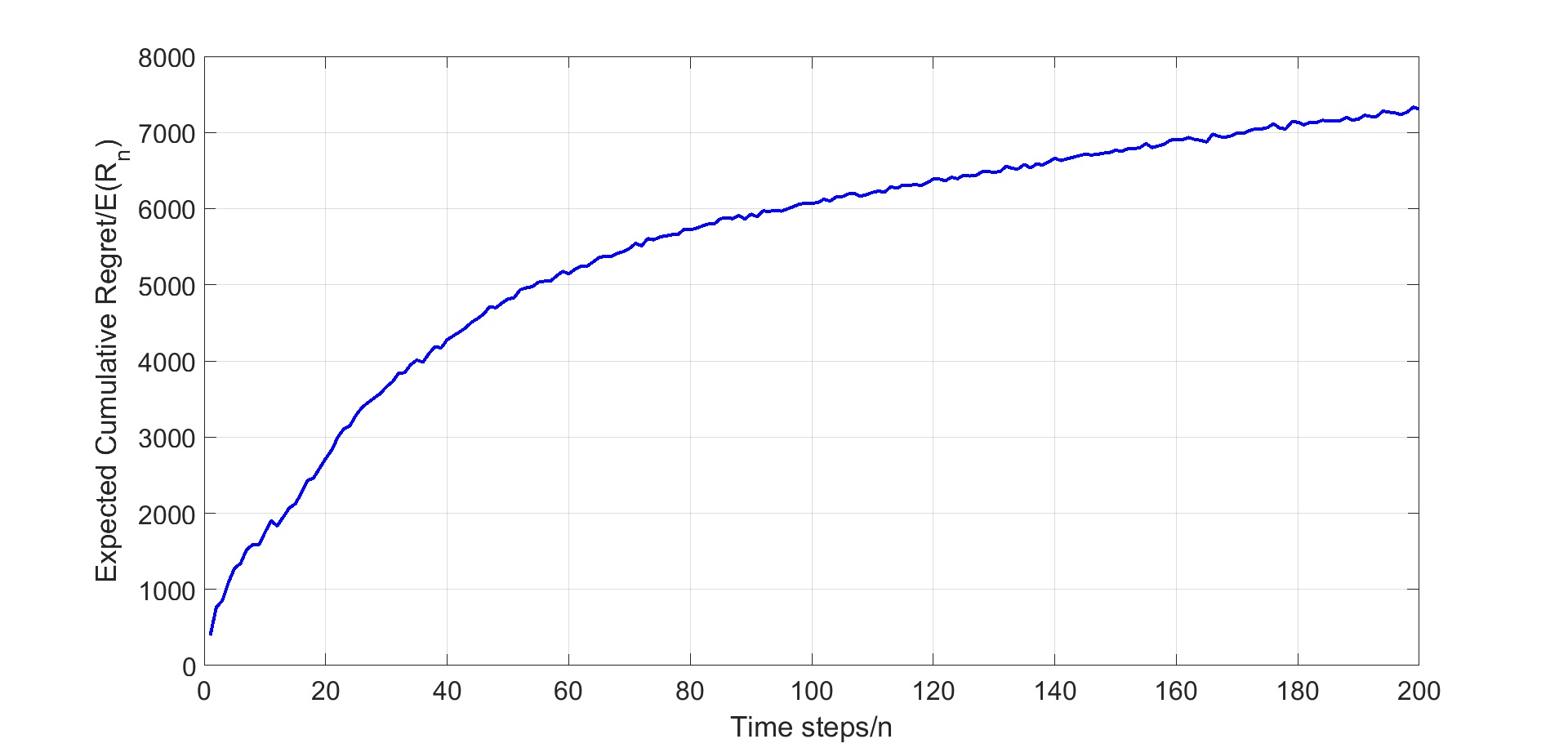}
 \caption{Expected cumulative regret $E(R_n)$  for the time varying DMAB with no noise. }
\label{Fig:EregretNoNoise}
\end{figure}

\begin{figure}[h!]
    \centering    \includegraphics[width=0.5\textwidth]{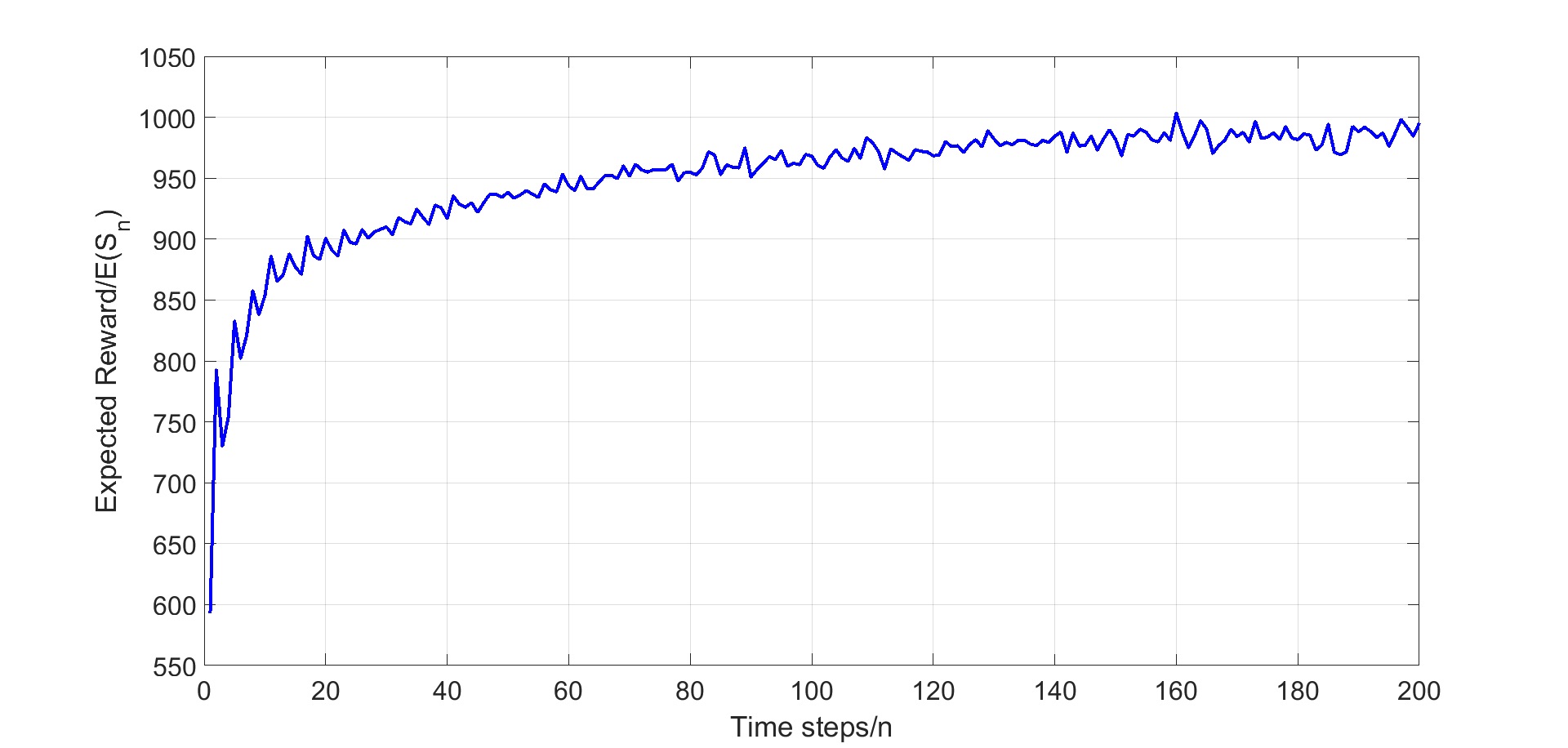}
 \caption{Expected reward $E(S_n)$ for the time varying DMAB with with uniform process noise.}
\label{Fig:ErewardNoise}
\end{figure}

\begin{figure}[h!]
    \centering    \includegraphics[width=0.5\textwidth]{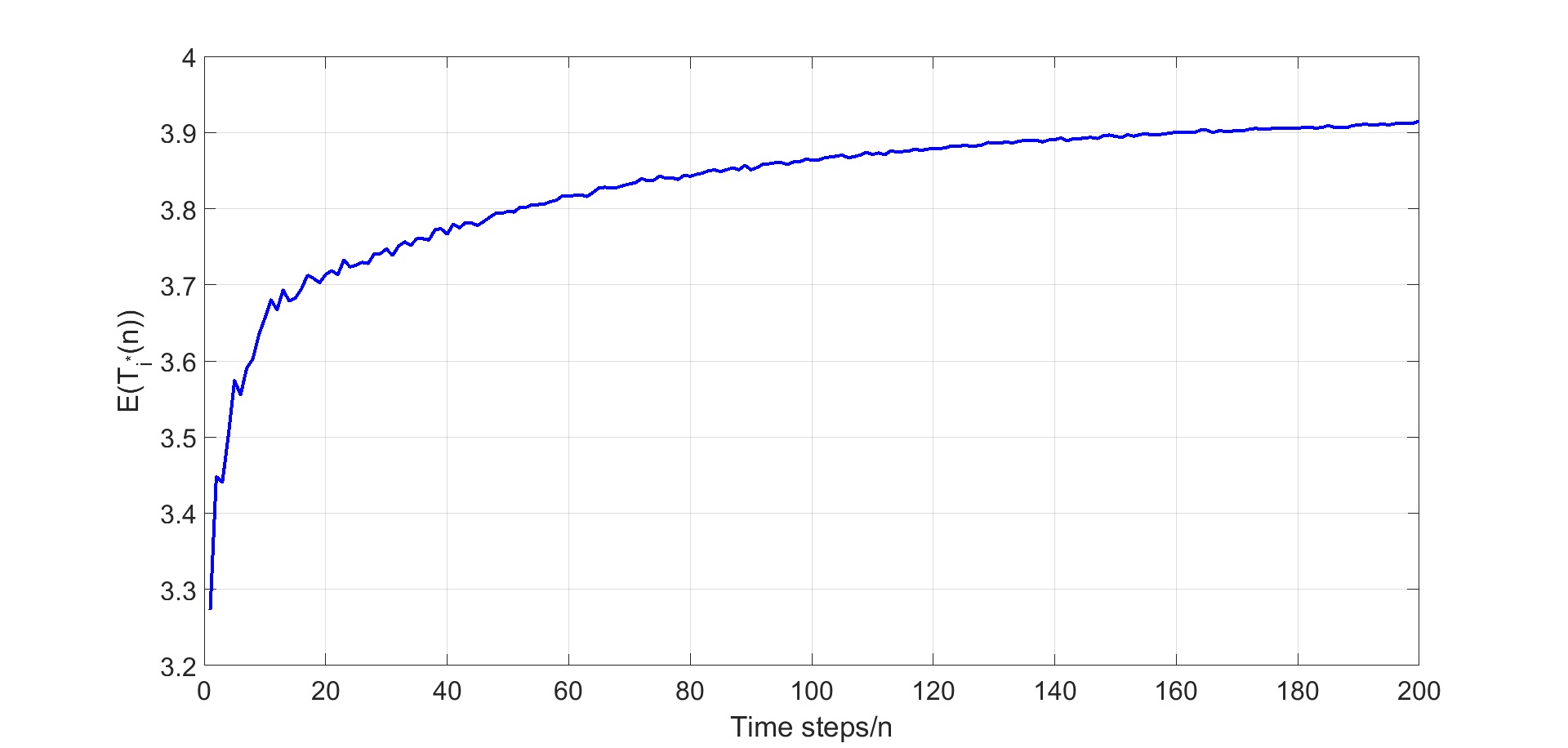}
 \caption{Expected number of times, $E(T_{i^*}(n))$, the optimal arm has been sampled  for the time varying DMAB with with uniform process  noise. }
\label{Fig:EToptNoise}
\end{figure}

\begin{figure}[h!]
    \centering    \includegraphics[width=0.5\textwidth]{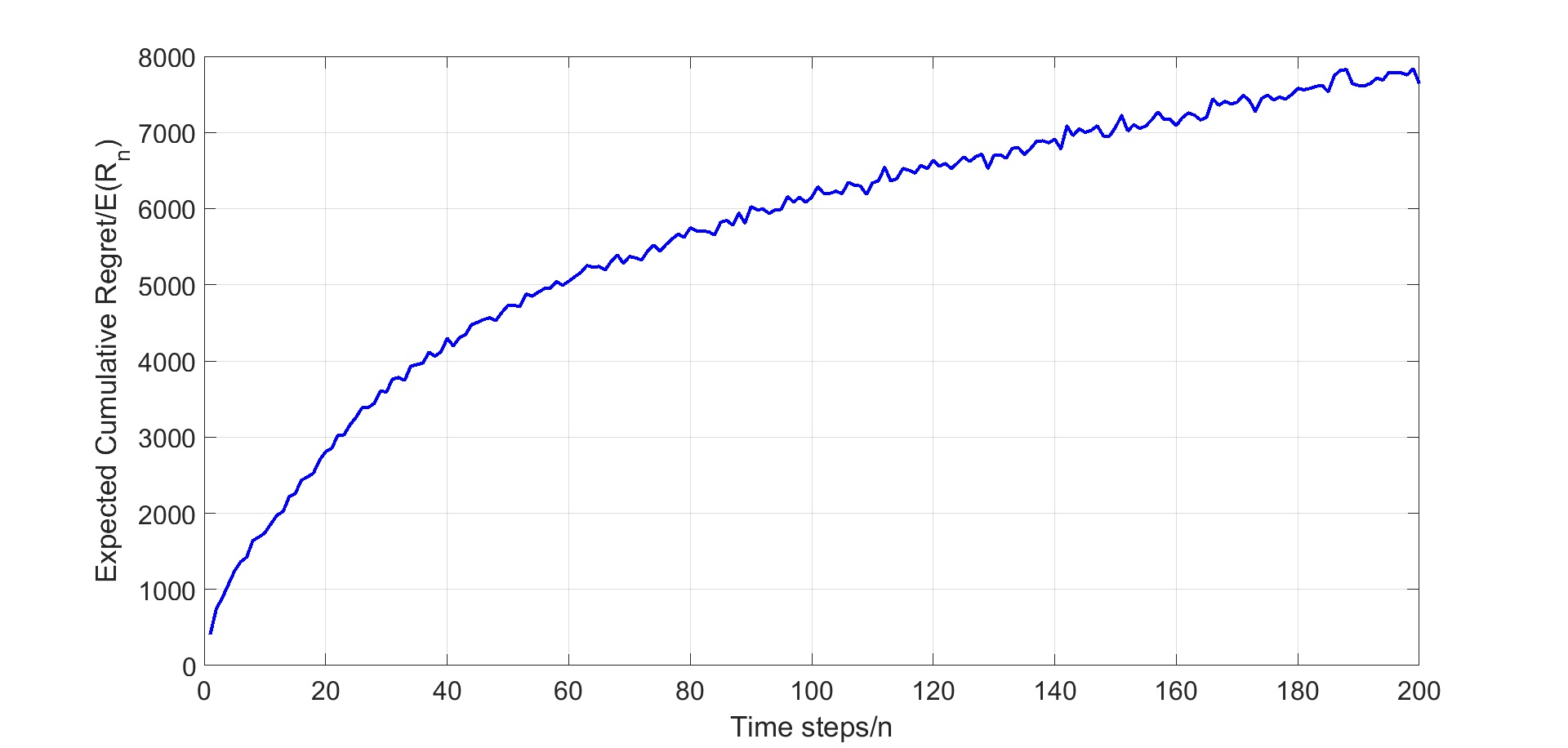}
 \caption{Expected cumulative regret $E(R_n)$  for the time varying DMAB with uniform process noise. }
\label{Fig:EregretNoise}
\end{figure}

The expected reward, $E(S_n)$, the expected number of times the optimal arm is sampled, $E(T_{i^*}^n)$, and the expected cumulative regret, $E(R_n)$, are plot against $n$ in Figurers \ref{Fig:ErewardNoNoise} -- \ref{Fig:EregretNoNoise} in the absence of process noise and in Figures \ref{Fig:ErewardNoise} -- \ref{Fig:EregretNoise} for uniformly distributed process noise. We observe that, as expected, the covariance of the regret and the reward increase as the number of time steps increase when the option dynamics are influenced by uncertainty. However since we only consider a finite time horizon they remain bounded during this horizon. Notice that yet the expected number of times the optimal arm is chosen behaves the same as when there is no process noise.

%%%%%%%%%%%%%%%

%%%%%%%%%%%%%%%%%%%%%%%%%%%%%%%%%%%%%%%%

%%%%%%%%%%%%%%%%%%%%%%%%%%%%%%%%%%%%%%%%

\section{Conclusion}
This paper presents a novel unifying framework for modeling a wide class of Dynamic Multi-Armed Bandit problems. It allows one to consider option unavailabilities and option correlations in a single setting. The class of problems is characterized by situations where the reward for each option depends uncertainly on a multidimensional parameter that evolves according to a linear stochastic dynamic system that captures the internal and hidden collective behavior of the dynamically changing options. The dynamic system is assumed to satisfy certain boundedness conditions. For this class of problems we show that the combination of any Upper Confidence Bound type algorithm with any efficient estimator guarantees that the expected cumulative regret is bounded above by a logarithmic function of the time steps. We provide a novel practically significant example to demonstrate these ideas.
%%%%%%%%%%%%%%%%%%%%%%%%%%%%%%

\begin{appendix}
\begin{proofoftheorem}
In the following we will proceed to prove the above Theorem \ref{Theom:DynamicRegret} by closely following the proof provided in \cite{Auer,Reverdy}.

Let $C_i^t\triangleq \sqrt{\frac{\Psi(t)}{T_i(t)}}$.
Then for $i\neq i^*$ and $l\geq 1$
\begin{align*}
E(T_i(n))&=\sum_{t=0}^{n-1}\mathcal{P}({\{\varphi_{t}=i\}})\leq l+\sum_{t=l}^{n-1}\mathcal{P}({\{\varphi_{t}=i\}})\\
&\leq l+\sum_{t=l}^{n-1}\mathcal{P}(\{Q_{i^*}^t< {Q^t_{i}}\}).
\end{align*}
Let
\begin{align*}
\mathcal{A}_i^t&\triangleq\{\widehat{X}_{i^*}^t+C_{i^*}^t\geq \widehat{\mu}^t_{i^*}\},\\
\mathcal{B}_i^t&\triangleq\{\widehat{\mu}^t_{i^*}\geq \widehat{\mu}^t_{i}+2{C^t_{i}}\},\\
\mathcal{C}_i^t&\triangleq\{\widehat{\mu}^t_{i}+2{C_{i}}^t\geq\widehat{X}_{i}^t+{C^t_{i}}\}\\
\mathcal{D}_i^t&\triangleq\{\gamma^t_{i^*}\neq 0\}
\end{align*} 

Then we have,
\begin{align*}
\{\mathcal{A}_i^t\cap \mathcal{B}_i^t \cap \mathcal{C}_i^t\}\cap \mathcal{D}_i^t\} \subseteq \{Q_{i^*}^t\geq{Q_{i}}^t\}
\end{align*}
Therefore,
\begin{align*}
\{Q_{i^*}^t< {Q^t_{i}}\} \subseteq {\bar{\mathcal{A}}_i^t}\cup {\bar{\mathcal{B}}_i^t} \cup {\bar{\mathcal{C}}_i^t}\cup {\bar{\mathcal{D}}_i^t}
\end{align*}
From above equations we have,
\begin{align*}
\mathcal{P}(\{Q_{i^*}^t< {Q^t_{i}}\})&\leq \mathcal{P}(\{\widehat{X}_{i^*}^t+C^t_{i^*}<\widehat{\mu}^t_{i^*}\})\\
&\:\:\:\:+\mathcal{P}(\{\widehat{\mu}^t_{i^*}< \widehat{\mu}^t_{i}+2{C^t_{i}}\})\\
&\:\:\:\:+\mathcal{P}(\{\widehat{X}_{i}^t-C^t_{i}>\widehat{\mu}^t_{i}\})\\
&\:\:\:\:+\mathbb{I}_{\{\gamma^t_{i^*}= 0\}}.
\end{align*}

Note that the conditions (\ref{eq:TailProbBnd1}) and (\ref{eq:TailProbBnd2}) of the tail probabilities of the distribution of the estimate $\widehat{X}_i^t$ gives us that,
\begin{align*}
\mathcal{P}(\{\widehat{X}_{i^*}^t+C_{i^*}^t<\widehat{\mu}^t_{i^*}\})&\leq\:\:\:\:\: \frac{\nu\,\log{t}}{\exp\left(2\kappa \sigma^2\Psi(t)\right)},\\
\mathcal{P}(\{\widehat{X}_{i}^t-C_{i}^t>\widehat{\mu}^t_{i}\})
&\leq\:\:\:\:\: \frac{\nu\,\log{t}}{\exp\left(2\kappa \sigma^2\Psi(t)\right)},
\end{align*}
and hence that
\begin{align*}
\mathcal{P}(\{Q_{i^*}^t< {Q^t_{i}}\})&\leq \mathbb{I}_{\{\gamma^t_{i^*}= 0\}}+\mathcal{P}(\bar{B}^t_i)+\frac{2\nu\,\log{t}}{\exp\left(2\kappa\sigma^2 \Psi(t)\right)}.
\end{align*}

From condition (\ref{eq:LogBndAvailability}) we have
$\sum_{j=2}^t\mathbb{I}_{\{\gamma^j_{i^*} =0\}}\leq \gamma \log{t}$. Thus
\begin{align*}
E(T_i(n))&\leq l+\gamma \log{n}+\sum_{t=l}^{n-1}\mathcal{P}(\bar{B}^t_i)+\sum_{t=l}^{n-1}\frac{2\nu\,\log{t}}{\exp\left(2\kappa \sigma^2\Psi(t)\right)}.
\end{align*}

%%%%%%%%%%%%%%%%%%%
Let us proceed to find an upper bound for $\sum_{t=1}^n\mathcal{P}(\bar{B}^t_i)$.
Since $\bar{B}^t_i=\{\widehat{\mu}^t_{i^*}< \widehat{\mu}_i^t+2{C^t_{i}}\}$.
Let $\Delta^t_{i}=\widehat{\mu}^t_{i^*}-\widehat{\mu}_i^t$. Then if $\bar{B}^t_i$ is true then,
\begin{align*}
\frac{\Delta^t_{i}}{2}&<\sigma\sqrt{\frac{\Psi(t)}{T_i(t)}},
\end{align*}
where the last inequality follows from (\ref{eq:UCBQ}).
Since $0<{\Delta_i}<\Delta^t_{i}$, thus we have that if $\bar{B}^t_i$ is true then 
\begin{align*}
T_i(t)&<\frac{4\sigma^2}{{\Delta^2_i}}\,\Psi(t),
\end{align*}
is true.
Thus for sufficiently large $t$
\begin{align*}
&\bar{B}^t_i\subseteq \left\{T_i(t)<\frac{4\sigma^2}{{\Delta^2_i}}\,\Psi(t)\right\},\\
&\mathcal{P}\left(\{\bar{B}^t_i\}\right)\leq \mathcal{P}\left(\left\{T_i(t)<\frac{4\sigma^2}{{\Delta^2_i}}\,\Psi(t)\right\}\right).
\end{align*}

Thus $\mathcal{P}\left(\bar{B}^t_i\right)\neq 0$ only if
\begin{align*}
T_i(t)&<\frac{4\sigma^2}{{\Delta^2_i}}\,\Psi(t).
\end{align*}
Thus we have
\begin{align*}
\sum_{t=l}^{n-1}\mathcal{P}(\bar{B}^t_i)&=\sum_{t=l}^{\tilde{t}}\mathcal{P}(\bar{B}^t_i)\leq \frac{4\sigma^2}{{\Delta^2_i}}\,\Psi(n),
\end{align*}
and hence
\begin{align*}
E(T_i(n))\leq l+\gamma \log{n}+\frac{4\sigma^2}{{\Delta^2_i}}\,\Psi(n) +\sum_{t=l}^{n-1}\frac{\nu\,\log{t}}{\exp\left(2\kappa \sigma^2\Psi(t)\right)}.
\end{align*}

If $\alpha \log{t}\leq \Psi(t) \leq \beta \log{t}$ then
\begin{align*}
E(T_i(n))\leq l+\gamma\log{n}+\frac{4\sigma^2\beta}{{\Delta^2_i}}\,\log n +\sum_{t=l}^{n-1}\frac{\nu\,\log{t}}{\exp\left(2\kappa\sigma^2 \alpha \log{t}\right)}.
\end{align*}
The series on the right converges as long as $\alpha>3/(2\kappa \sigma^2)$.
Thus from  (\ref{eq:DynRegret}) we have that the cumulative expected regret satisfies:
\begin{align*}
R_n\leq c_1+ \bar{\Delta}\sum_{i\neq i^*}^k\left(\gamma+\frac{4 \sigma^2\beta}{\Delta_i^2}\right)\,\log{n} ,
\end{align*}
where
\[
c_1=k\bar{\Delta}\nu\,\left(\frac{\log{2}}{2^{2\kappa\sigma^2 \alpha}}+\int_2^{n-1}\frac{1}{t^{2\kappa \sigma^2\alpha-1}}dt\right)
\]
The integral on the right converges as $n\to \infty$ if $\alpha>3/(2\kappa\sigma^2)$. Thus completing the proof of Theorem \ref{Theom:DynamicRegret}.
\end{proofoftheorem}

\end{appendix}
\bibliographystyle{IEEEtran}
\bibliography{DynamicBandit}

\newpage
\end{document}